\documentclass[letterpaper, 10 pt, conference]{ieeeconf}  

\IEEEoverridecommandlockouts                              

\usepackage{xcolor}
\usepackage{placeins}
\usepackage{graphics} 
\usepackage{epsfig} 
\usepackage{mathptmx} 
\usepackage{times} 
\usepackage{multirow}
\usepackage{amsmath} 
\usepackage{amssymb}  
\usepackage{algorithm,algcompatible}
\usepackage{siunitx}
\usepackage{subcaption}
\usepackage[hidelinks]{hyperref}
\algnewcommand\INPUT{\item[\textbf{Input:}]}%
\algnewcommand\OUTPUT{\item[\textbf{Output:}]}%
\newcommand{\DECO}[0] {(DE)\textsuperscript{2}CO}
\title{\LARGE \bf
(DE)$^2$CO: Deep Depth Colorization
}

\author{F. M. Carlucci*, P. Russo*, and B. Caputo$^{1}$
\thanks{This work was partially supported by the ERC grant 637076 - RoboExNovo (F.M.C., P. R., B.C.), and the CHIST-ERA project ALOOF (B.C, F. M. C., P. R.).}
\thanks{$*$ Equal contribution}
\thanks{$^{1}$ All authors are at the Department of Computer, Control, and Management Engineering  Antonio Ruberti at Sapienza Rome University, Italy and at the Italian Institute of Technology, VANDAL Laboratory, Milan, Italy
        {\tt\small \{ fabio.carlucci,paolo.russo, barbara.caputo \} @iit.it}}
}

\begin{document}

\maketitle
\thispagestyle{empty}
\pagestyle{empty}

\begin{abstract}
The ability to classify objects is fundamental for robots. Besides knowledge about their visual appearance, captured by the RGB channel, robots heavily need also depth information to make sense of the world. While the use of deep networks on RGB robot images has benefited from the plethora of results obtained on databases like ImageNet, 
using convnets on depth images requires mapping them into three dimensional channels. This transfer learning procedure makes them processable by pre-trained deep architectures. Current mappings are based on heuristic assumptions over pre-processing steps and on what depth properties  should be most preserved, resulting often in cumbersome data visualizations, and in sub-optimal performance in terms of generality and
recognition results. Here we take an alternative route and we attempt instead to \emph{learn}  an optimal colorization mapping for any given pre-trained architecture, using as training data a reference RGB-D database. We propose a deep network architecture, exploiting the residual  paradigm,  that learns how to map depth data to three channel images. 
A qualitative analysis of the images obtained with this approach clearly indicates that learning the optimal mapping  preserves the richness of depth information better than
current hand-crafted approaches. 
Experiments on the Washington, JHUIT-50 and BigBIRD  public benchmark databases, using CaffeNet, VGG-16, GoogleNet, and ResNet50  clearly showcase the power of our approach, with gains in  performance of up to $16\%$ compared to state of the art competitors on the depth channel only, leading to top performances when dealing with RGB-D data.

\end{abstract}

\section{INTRODUCTION}

Robots need to recognize what they see around them to be able to act and interact with it. Recognition must be carried out in the RGB domain, capturing  mostly the visual appearance of things related to their reflectance properties, as well as in the depth domain, providing information about the shape and silhouette of objects and supporting both recognition and interaction with items.  The current mainstream state of the art approaches for object recognition are based on Convolutional Neural Networks (CNNs, ~\cite{NIPS1989_293-CNN}), which use end-to-end architectures achieving feature learning and classification at the same time. Some notable advantages of these networks are their ability to reach much higher accuracies on basically any visual recognition problem, compared to what would be achievable with heuristic methods; their being domain-independent, and their conceptual simplicity. Despite these advantages, they also present some limitations, such as high computational cost, long training time and the demand for large datasets, among others.

This last issue has so far proved crucial in the attempts to leverage over the spectacular success of CNNs over RGB-based object categorization \cite{googlenet,krizhevsky2012imagenet} in the depth domain. Being CNNs data-hungry algorithms, the availability of very large scale annotated data collections is crucial for their success, and architectures trained over ImageNet \cite{deng2009imagenet} are the cornerstone of the vast majority of CNN-based recognition methods. Besides the notable exception of \cite{carlucci2016deep}, the mainstream approach for using CNNs on depth-based object classification has been through transfer learning, in the form of a mapping able to  make the depth input channel compatible with the data distribution 
expected by RGB architectures.
Following
recent efforts in transfer learning~\cite{Transfer1,Transfer2,Transfer3} that made it possible to use depth data with CNN pre-trained on a database of a different modality,
several authors proposed hand-crafted mappings to colorize depth data, obtaining impressive improvements in classification 
over the Washington \cite{washington} database, that has become the golden reference benchmark in this field ~\cite{Schwarz,eitel2015multimodal}.

\begin{figure*}[!ht]
\centering
\includegraphics[width=0.9\textwidth]{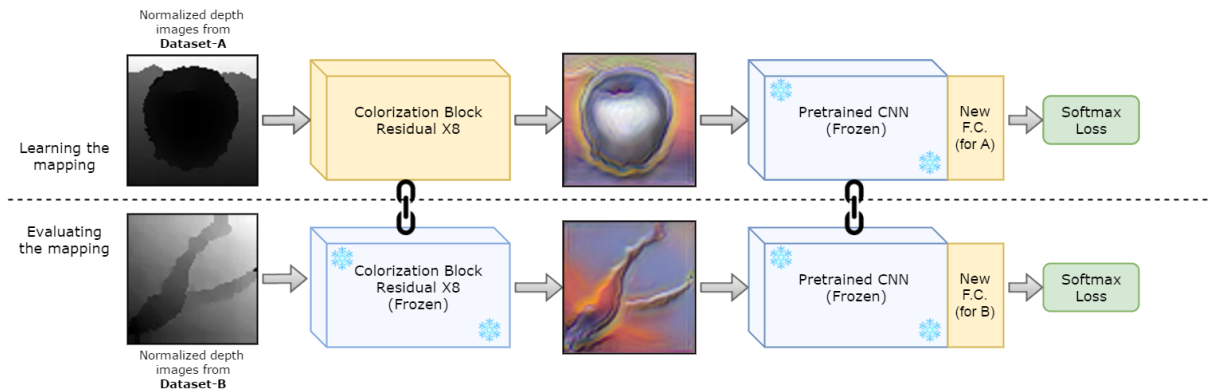}
\caption{The \DECO{} pipeline consists of two phases. First, we learn the mapping, from depth to color, maximizing the discrimination capabilities of a network pre trained on ImageNet. In this step the network is frozen and we are only learning the mapping and the final layer. 
We then evaluate the colorization on a \emph{different} depth dataset: here we also freeze the colorization network and only train a new final layer for the testbed dataset.}
\label{fig:deco_arch}
\vspace{-0.2cm}
\end{figure*}

We argue that this strategy is sub-optimal. By hand-crafting the mapping for the depth data colorization, one has to make strong assumptions on what information, and up to which extent, should be preserved in the transfer learning
towards the RGB modality. 
While some choices might be valid for some classes of problems and settings, it is questionable whether the family of algorithms based on this approach can provide results  combining high recognition accuracies with robustness across different settings and databases. Inspired by recent works on colorization of gray-scale photographs \cite{IizukaSIGGRAPH2016,larsson2016learning,cheng2015deep}, we tackle the problem by exploiting the power of end-to-end convolutional networks, proposing
 a deep depth colorization architecture able to learn the optimal 
 transfer learning from depth to RGB
  for any given pre-trained convnet. Our deep colorization network takes advantage of the residual approach \cite{ResNet}, 
learning how to map between the two modalities
by leveraging over a reference database (Figure \ref{fig:deco_arch}, top), for any given architecture. After this training stage, the colorization network can be added on top of its reference pre-trained architecture, for any object classification task  (Figure \ref{fig:deco_arch}, bottom).
We call our network \DECO: DEep DEpth COlorization. 

We assess the performance of \DECO\ in several ways. A first qualitative analysis, comparing the colorized depth images obtained by \DECO\  and by other state of the art hand-crafted approaches, gives intuitive insights on the advantages brought by learning the mapping as opposed to choosing it, over several databases. We further deepen this analysis with an experimental evaluation of our and other existing transfer learning methods on the depth channel only, using four different deep architectures and three different public databases, with and without fine-tuning. Finally, we tackle the RGB-D object recognition problem, combining  \DECO\  with off-the shelf state of the art RGB deep networks, benchmarking it against the current state of the art in the field. For all these experiments, results clearly support the value of our algorithm. 
All the \DECO\  modules, for all architectures employed in this paper, are available at \url{https://github.com/fmcarlucci/de2co}.


\section{RELATED WORK}
\label{related}
\label{rel-work}
Since 2012's  AlexNet \cite{krizhevsky2012imagenet} spectacular success,
CNNs have become the dominant learning paradigm in visual recognition. Several architectures have been proposed 
in recent years,
each bringing new flavors to the community.  Simonyan and Zisserman~\cite{VGG} investigated the effect of increasing the network depth.
GoogLeNet \cite{googlenet} also increased the depth and width of the network while restraining the computational budget, with a dramatic reduction in the number of parameters. 
 He et al. \cite{ResNet} proposed a residual learning approach using a batch normalization layer and special skip connections for training deeper architectures, showing an impressive success in ILSVRC2015. 
 All these architectures will be used in this work, to assess its generality.

Lately, several authors attempted to take advantage of pre-trained CNNs to perform RGB-D detection and recognition.
Colorization of depth images can be seen as a transfer learning process across modalities, and several works explored this avenue within the deep learning framework.
In the context of RGB-D object detection, a recent stream of works explicitly addressed cross modal transfer learning through sharing of weights across architectures \cite{hoffman2016icra}, \cite{hoffman2016learning} and \cite{Gupta_2016_CVPR}. This last work is conceptually close to our approach, as it proposes to learn how to transfer RGB extracted information to the Depth domain through distillation \cite{hinton2015distilling}. While  \cite{Gupta_2016_CVPR} has proved very successful in the object detection realm, it presents some constraints that might potentially be problematic in object recognition, from the requirement of paired RGB-D images, to detailed data preprocessing and preparation for training. As opposed to this, our algorithm
does not require explicit pairing of images in the two modalities, can be applied successfully on raw pixel data and does not require other data preparation for training.

Within the RGB-D classification literature,
 ~\cite{Bo} converts the depth map to surface normals and then re-interprets it as RGB values, while
Aekerberg et al. \cite{aakerberg2017depth} builds on this approach and suggests an effective preprocessing pipeline to increase performance.
A new method has also been proposed by Gupta, Saurabh, et al. \cite{gupta2014learning} 
: HHA is a mapping where one channel encodes the horizontal disparity, one the height above ground and the third the pixelwise angle between the surface normal and the gravity vector. 
Schwarz et al ~\cite{Behnke} proposed a  colorization pipeline where colors are assigned to the image pixels according to the distance of the vertexes of a rendered mesh to the center of the object. 
Besides the naive grayscale method, the rest of the mentioned colorization schemes are computationally expensive.  Eitel et al ~\cite{eitel2015multimodal} used a simple color mapping technique known as \textit{ColorJet},  showing this simple method to be competitive with more sophisticated approaches.


All these works, and many others \cite{zaki2016convolutional,carlucci2016deep}, make use of an ad-hoc mapping for converting depth images into three channels.
This conversion is vital as the dataset has to be compatible with the pre-trained CNN. Depth data is encoded as a 2D array where each element represents an approximate distance between the sensor and the object. Depth information is often depicted and stored as a single monochrome image.  Compared to regular RGB cameras, the depth resolution is relatively low, especially when the frame is cropped to focus on a particular object. 
While addressing this issue, we avoid heuristic choices in our approach and we rely instead on an end-to-end, residual based deep architecture to learn the optimal mapping for the cross modal knowledge transfer.


Most of works in object recognition, against whom we compare our method, 
are evaluated
on one single database, with Washington being the standard choice in the robot vision literature. This raises concerns about 
the generality of these methods, especially considering their hand-crafted nature. We circumvent this issue 
by evaluating \DECO\ on three different databases.


Our work is also related to the colorization of grayscale images using deep nets. 
Cheng et al ~\cite{cheng2015deep} proposed a colorization pipeline based on three different hand-designed feature extractors to determine the features from different levels of an input image.  Larsson et al ~\cite{larsson2016learning} used an architecture consisting of two parts. The first part is a fully convolutional version of VGG-16 used as  feature extractor, and the second part is a fully-connected layer with 1024 channels predicting the distributions of hue and the chroma for each pixel given its feature descriptors from the previous level. 
Iizuka et al ~\cite{IizukaSIGGRAPH2016} proposed an end-to-end network  able to learn global and local features,  exploiting the classification labels for better image colorization. Their architecture  consists of several networks followed by fusion layer for the colorization task.
Sun et al. \cite{sun2017weakly} propose to use large scale CAD rendered data to leverage depth information without using low level features or colorization. In Asif et al. \cite{asif2017rgb}, hierarchical
cascaded forests were used for computing grasp poses
and perform object classification, exploiting several different features like orientation angle maps, surface normals and depth information colored with \textit{Jet} method.
Our work differs from this last research thread in the specific architecture proposed, and in its main goal, as here we are interested in learning optimal mapping for categorization rather than for colorization of grayscale images.

\section{COLORIZATION OF DEPTH IMAGES}
\label{colorization}
Although depth and RGB are modalities with significant differences,
they also share enough similarities (edges, gradients, shapes)
to make it plausible that convolutional filters learned from RGB data could be re-used effectively for representing colorized depth images. The approach currently adopted in the literature consists of designing ad-hoc colorization algorithms, as revised in the previous section. We refer to these kind of approaches as \emph{hand-crafted depth colorization}. Specifically, we choose ColorJet 
\cite{eitel2015multimodal}, SurfaceNormals \cite{Bo} and \textit{SurfaceNormals++}  
\cite{aakerberg2017depth} as baselines against which we will assess our data driven approach because of their popularity and effectiveness.




In the rest of the section we first briefly summarize ColorJet  (section \ref{shallow-depth}), SurfaceNormals and SurfaceNormals++  (section \ref{shallow-depth-surfacenormals}). We then describe our deep approach to depth colorization (section \ref{de2co}). To the best of our knowledge, \DECO\ is the first deep colorization architecture applied successfully to depth images.


\subsection{Hand-Crafted Depth Colorization: ColorJet}
\label{shallow-depth}
ColorJet works by assigning different colors to different depth values. The original depth map is firstly normalized between 0-255 values. Then the colorization works by  mapping the lowest value to the blue channel and the highest value to the red channel. The value in the middle is mapped to green and the intermediate  values are arranged accordingly \cite{eitel2015multimodal}. The resulting image exploits the full RGB spectrum, with the intent of leveraging at best the filters learned by deep networks trained on very large scale RGB datasets like ImageNet. Although simple, the approach gave very strong results when tested on the Washington database, and when deployed on a robot platform. Still, ColorJet was not designed to create
realistic looking RGB images for the objects depicted in the original depth data (Figure \ref{fig:multiple_mappings}, bottom row). This  raises the question whether this mapping, although more effective than other methods presented in the literature, might be sub-optimal. In section \ref{de2co}  we will show that by fully embracing the end-to-end philosophy at the core of deep learning, it is indeed possible to achieve significantly higher recognition performances while at the same time producing more realistic colorized images.   

\subsection{Hand-Crafted Depth Colorization: SurfaceNormals(++)}
\label{shallow-depth-surfacenormals}
The SurfaceNormals mapping has been often used to convert depth images to RGB \cite{Bo,wang2016correlated,eitel2015multimodal}. The process is straightforward:  for each pixel in the original image the corresponding surface normal is computed as a normalized $3D$ vector, which is then treated as an RGB color. Due to the inherent noisiness of the depth channel, such a direct conversion results in noisy images in the color space. To address this issue, the mapping we call \textit{SurfaceNormals++} was introduced by Aakerberg \cite{aakerberg2017depth}: first, a recursive median filter is used to reconstruct missing depth values, subsequently a bilateral filter smooths the image to reduce noise, while preserving edges.
Next, surface normals are computed for each pixel in the depth image. Finally the image is sharpened using the unsharp mask filter, to increase contrast around edges and other
high-frequency components.

\subsection{Deep Depth Colorization: (DE)$^2$CO}
\label{de2co}
\DECO{}
consists of feeding the depth maps, normalized into grayscale images, to a \textit{colorization network} linked to a standard CNN architecture, pre-trained on ImageNet.

\begin{figure}
\centering
\includegraphics[width=0.4\textwidth]{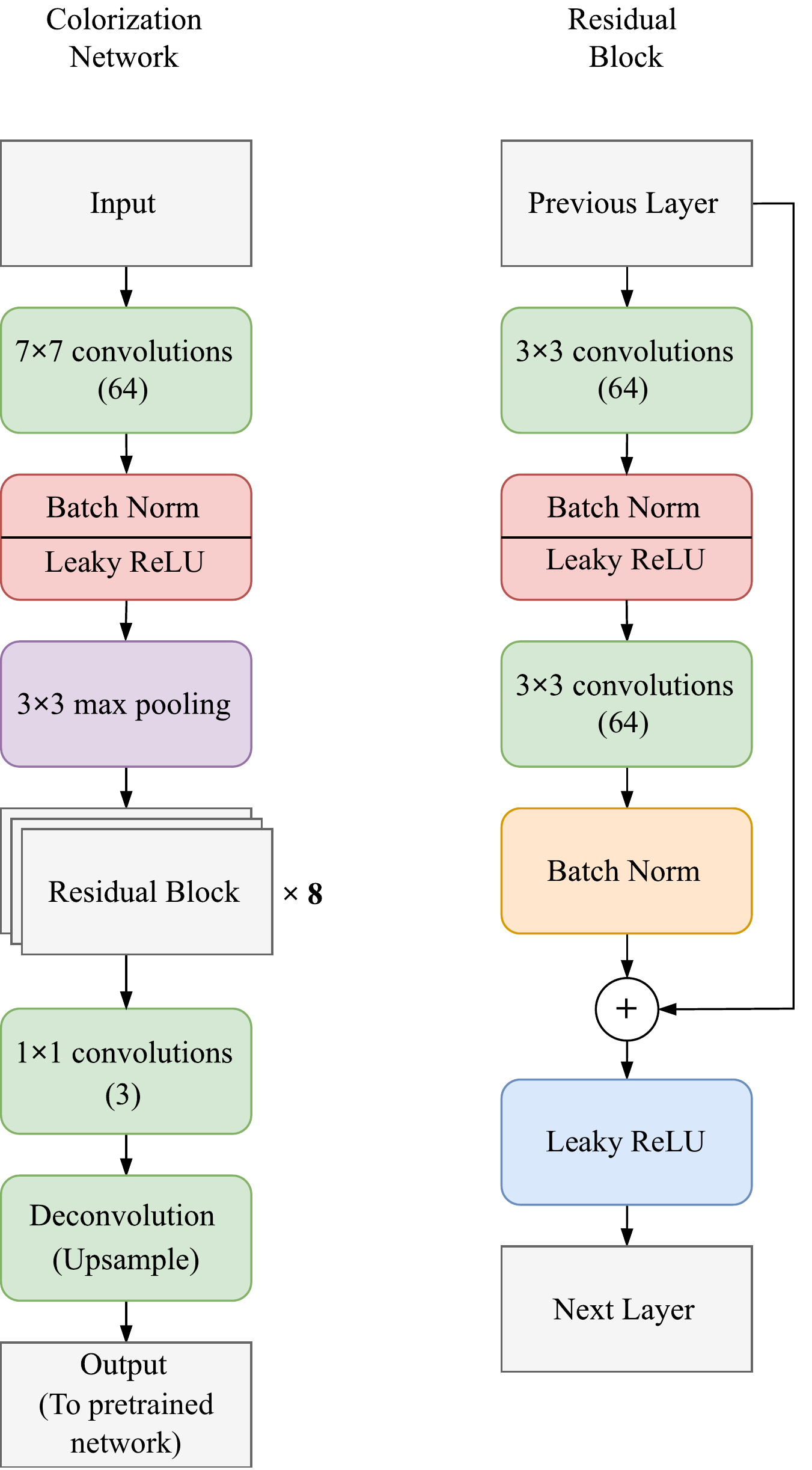}
\caption{Overview of the  \DECO\ colorization network. On the left, we show the overall architecture; on the right, we show details of the residual block.}
\label{fig:rescol}
\vspace{-0.6cm}
\end{figure}

Given the success of deep colorization networks from grayscale images, we first tested   existing architectures in this context \cite{zhang2016colorful}. Extensive experiments showed that while the visual appearance of the colorized images was very good, the recognition performances obtained when combining such network with pre-trained RGB architectures was not competitive. Inspired by the generator network in \cite{bousmalis2016unsupervised}, we propose here a \textit{residual} convolutional architecture (Figure \ref{fig:rescol}). 
By design \cite{ResNet}, this architecture is robust and allows for deeper training. This is helpful here, as \DECO\ requires stacking together two networks, which for not very deep architectures might lead to vanishing gradient issues.
Furthermore, residual blocks works at pixel level \cite{bousmalis2016unsupervised} helping to preserve locality.

Our architecture works as follows: the 1x228x228 input depth map, reduced to 64x57x57 size by a conv\&pool layer, passes through a sequence of 8 residual blocks, composed by 2 small convolutions with batch normalization layers and leakyRelu as non linearities. 
The last residual block output is convolved by a three features convolution to form the 3 channels image output. Its resolution is brought back to 228x228 by a \textit{deconvolution} (upsampling) layer. 

Our whole system for object recognition in the depth domain using  deep networks pre-trained over RGB images can be summarized as follows:
the entire network, composed by  \DECO\  and the classification network of choice, is trained on an annotated reference  depth image dataset.  The weights of the chosen classification network are kept frozen in their pre-trained state, as the only layer that needs to be retrained is the last fully connected layer connected to the softmax layer.  Meanwhile, the weights of   \DECO\  are updated until convergence.

\begin{figure*}
\centering
\includegraphics[trim=0 0 0 0,clip,width=0.85\textwidth]{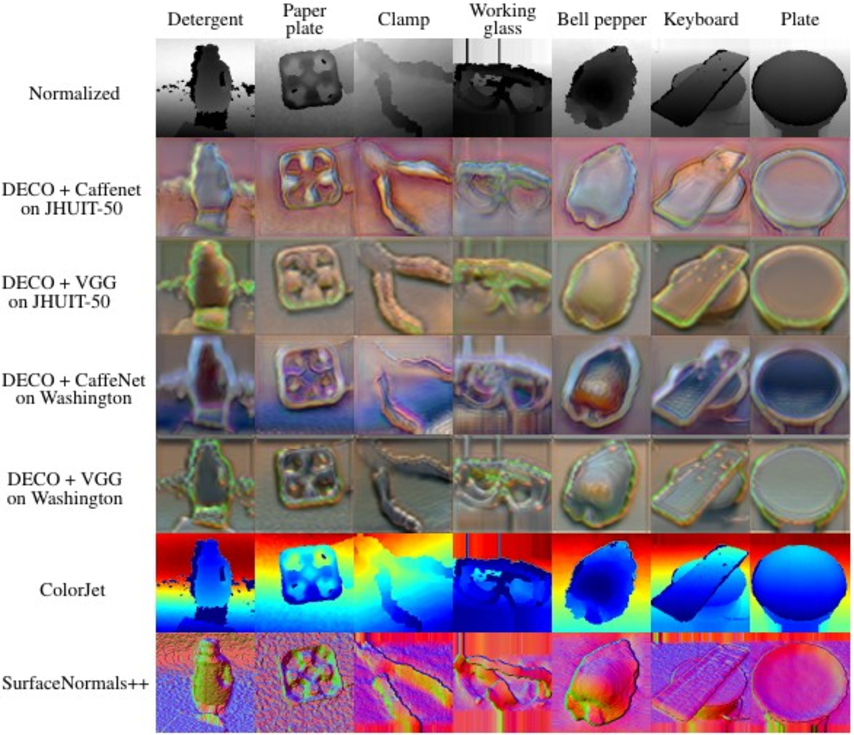}
\caption{ {(DE)$^2$CO} colorizations applied on different objects, taken from\cite{jhuit,washington,singh2014bigbird}. Top row shows the depth maps mapped to grayscale. From the second to the fourth row, we show the corresponding {(DE)$^2$CO} colorizations learned on different settings. Fifth row shows \textit{ColorJet} views \cite{eitel2015multimodal}, while the last row shows the surface normals mapping.\cite{aakerberg2017depth} \textit{SurfaceNormals++}. These images showcase \DECO{}'s ability 
to emphasize the object's shape and to capture its salient features. 
}
\label{fig:multiple_mappings}
\vspace{-0.5cm}
\end{figure*}

After this step, the depth colorization network has learned the mapping that maximizes the classification accuracy on the reference training dataset. 
It can now be used to colorize \emph{any} depth image, from any data collection. 
Figure \ref{fig:multiple_mappings}, top rows, shows exemplar images colorized with \DECO\ trained over different reference databases, in combination with two different architectures (CaffeNet, an implementation variant of AlexNet, and VGG-16 \cite{VGG}). We see that, compared to the images obtained with ColorJet and SurfaceNormal++, our colorization technique emphasizes the objects contours and their salient features while flatting the object background, while the other methods introduce either high frequency noise (SurfaceNormals++) or emphasize background gradient instead of focusing mainly on the objects (ColorJet).    
In the next section we will show how this qualitative advantage translates also into a numerical advantage, i.e.  how learning \DECO\ on one dataset and performing depth-based object recognition on another
leads to a very significant increase in performance on several settings, compared to hand-crafted color mappings.





\label{method}

\section{EXPERIMENTS}
\label{experiments}
\label{setup}
We evaluated our colorization scheme on three main settings: an ablation study of how different depth mappings perform when the network weights are kept frozen (section \ref{ablation}), a comparison of depth performance with network finetuning (section \ref{fine-tuning}) and finally an assessment of \DECO\ when used in RGB-D object recognition tasks (section \ref{rgb-d}). Before reporting on our findings, we illustrate the databases and baselines we used  (section \ref{exper-setup}).

\subsection{Experimental Setup}
\label{exper-setup}

\noindent
\textbf{Databases} We considered three datasets: the Washington RGB-D \cite{washington}, the JHUIT-50 \cite{jhuit} and the BigBIRD \cite{singh2014bigbird} object datasets, which are the main public datasets for RGB-D object recognition. The first consists of $41,877$ RGB-D images organized into $300$ instances divided in $51$ classes. 
We performed experiments on the object categorization setting, where we followed the evaluation protocol defined in \cite{washington}. The JHUIT-50  is a challenging recent dataset that focuses on the problem of fine-grained classification.  It contains $50$ object instances, often very similar with each other (e.g. 9 different kinds of screwdrivers). As such, it presents different recognition challenges compared to the Washington database. Here we followed the evaluation procedure defined in \cite{jhuit}.
BigBIRD is the biggest of the datasets we considered: it contains $121$ object instances and $75.000$ images. Unfortunately, it is an extremely unforgiving dataset for evaluating depth features: many objects are extremely similar, and many are boxes, which are indistinguishable without texture information. To partially mitigate this, we grouped together all classes annotated with the same first word:
for example \textit{nutrigrain apple cinnamon} and \textit{nutrigrain blueberry} were grouped into \textit{nutrigrain}. With this procedure, we reduced the number of classes to 61 (while keeping all of the samples). 
As items are quite small
, we used the object masks provided by \cite{singh2014bigbird} to crop around the object. Evaluation-wise, we followed the  protocol defined in \cite{jhuit}.

\noindent
\textbf{Hand-crafted Mappings}
According to previous works \cite{eitel2015multimodal,aakerberg2017depth}, the two most effective mappings are \textit{ColorJet} \cite{eitel2015multimodal} and \textit{SurfaceNormals} \cite{Bo,aakerberg2017depth}.
For ColorJet we normalized the data between $0$ and $255$ and then applied the mapping using the OpenCV libraries\footnote{"COLORMAP\_JET" from \textit{http://opencv.org/}}. 
For the SurfaceNormals mapping we considered two versions: the straightforward conversion of the depthmap to surface normals \cite{Bo} and the enhanced version \textit{SurfaceNormals++}\cite{aakerberg2017depth} which uses extensive pre-processing and generally performs better\footnote{The authors graciously gave us their code for our experiments.}.

\subsection{Ablation Study}
\label{ablation}
In this setting we compared our \DECO\ method against hand crafted mappings, using pre-trained networks as feature extractors and only retraining the last classification layer.
We did this on the three datasets described above,  over four architectures:
 CaffeNet (a slight variant of the AlexNet \cite{NIPS2012-AlexNet}), VGG16 \cite{VGG} and GoogleNet \cite{googlenet} were chosen because of their popularity within the robot vision community.  We also tested the recent ResNet50 \cite{ResNet}, which 
although not currently very used in the robotics domain, has some promising properties.
In all cases we considered models pretrained on ImageNet \cite{deng2009imagenet}, which we retrieved from Caffe's \textit{Model Zoo}\footnote{\textit{https://github.com/BVLC/caffe/wiki/Model-Zoo}}.

Training \DECO\ means minimizing the multinomial logistic loss of a network trained on RGB images. This means that our  network is attached between the depth images and the pre-trained network, of which we freeze the weights of all but the last layer, which are relearned from scratch (see Figure \ref{fig:deco_arch}).
We trained each network-dataset combination for $50$ epochs using the Nesterov solver \cite{nesterov1983method} and $0.007$ starting learning rate (which is stepped down after $45\%$). During this phase, we  used the whole  \textit{source} datasets, leaving aside only $10\%$ of the samples for validation purposes.

When the dataset on which we train the colorizer is different from the test one, we simply retrain the new final layer (freezing all the rest) for the new classes.

Effectively we are using the pre-trained networks as feature extractors, as done in \cite{Behnke,eitel2015multimodal,zaki2016convolutional} and many others; for a performance analysis in the case of network finetuning we refer to paragraph \ref{fine-tuning}.  In this setting we used the Nesterov (for Washington and JHUIT-50) and ADAM (for BigBIRD) solvers. As we were only training the last fully connected layer, we learned  a small handful of parameters with a very low risk of overfitting.

\begin{figure*}[!ht]
\centering
\includegraphics[width=0.92\textwidth]{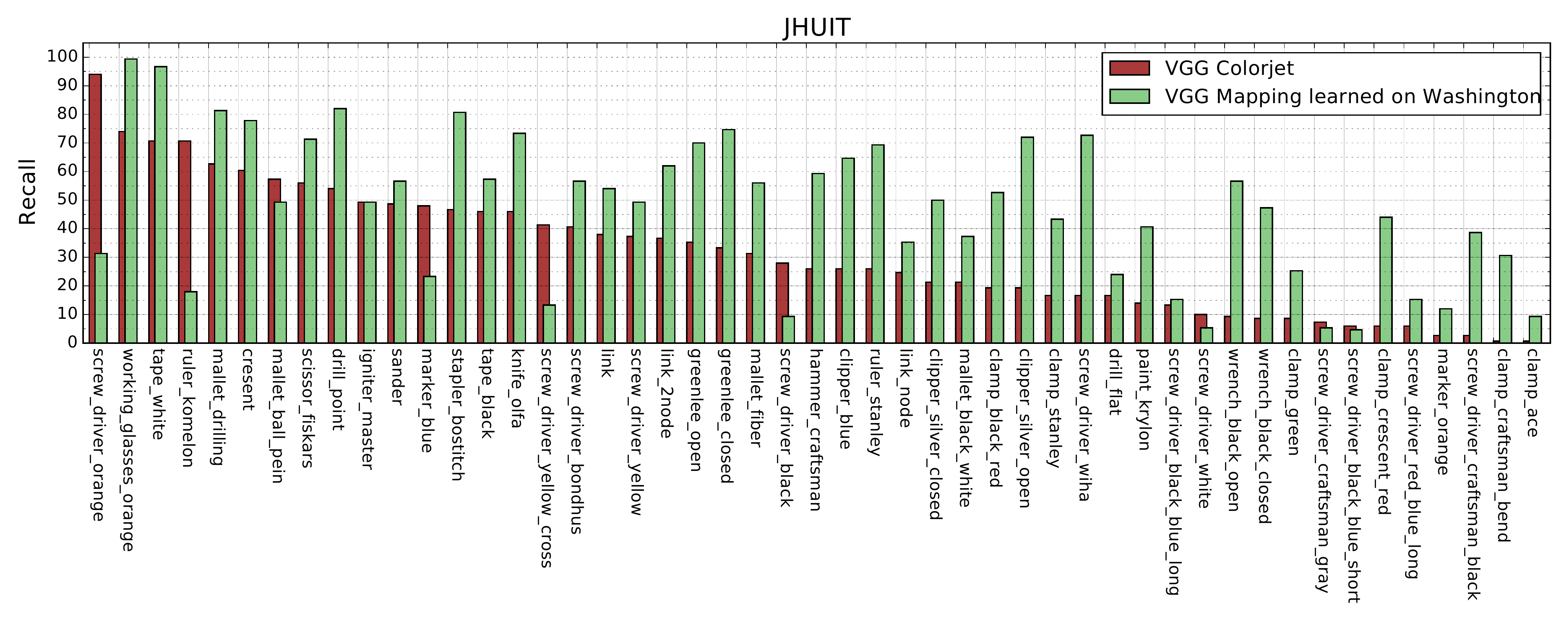}
\caption{
Per class recall on JHUIT-50, using VGG, with \DECO\ learned from Washington. Recalls per class are sorted in decreasing order, according to ColorJet performance. In this setting, 
\DECO{}, while generally performing better, seems to focus on different perceptual properties and is thus, compared with the baseline, better at some classes rather than others. }
\label{fig:jhuit_recall}
\vspace{-0.2cm}
\end{figure*}

Table \ref{table:comparison} reports the results from the ablation
while Figure \ref{fig:jhuit_recall} focuses on the class recall for a specific experiment.
For every architecture, we report the results obtained using ColorJet, SurfaceNormals (plain and enhanced) and \DECO\ learned on a reference database between Washington or JHUIT-50, and \DECO\ learned on the combination of Washington and JHUIT-50. For the CaffeNet and VGG networks we also present results on simple grayscale images.
We attempted also to learn \DECO\ from BigBIRD alone, and in combination with one (or both) of the other two databases. Results on BigBIRD only were disappointing, and results with/without adding it to the other two databases did not change the overall performance. We interpret this result as caused by the relatively small variability of objects in BigBIRD with respect to depth, and for sake of readability we decided to omit  them in this work.


We see that, for all architectures and for all reference databases, \DECO\ achieves higher results. The difference goes from $+1.7\%$, obtained with CaffeNet on the Washington database, to the best of $+16.8\%$ for VGG16 on JHUIT-50.  JHUIT-50 is the testbed database where, regardless of the chosen architecture, \DECO\ achieves the strongest gains in performance compared to hand crafted mappings. Washington is, for all architectures, the database where hand crafted mappings perform best, with the combination Washington to CaffeNet being the most favorable to the shallow mapping. 
On average it appears the CaffeNet is the architecture that performs best on this datasets; still, it should be noted that we are using here all architectures as feature extractors rather than as classifiers. On this type of tasks, both ResNet and GoogLeNet-like networks are known to perform worse than CaffeNet \cite{azizpour2016factors}, hence our results are consistent with what reported in the literature.  
In Table \ref{table:DECO_ablation} we report a second ablation study performed on the width and depth of \DECO{} architecture. Starting from the standard \DECO{} made of 8 residual blocks with 64 filters for each convolutional layer (which we found to be the best all-around architecture),
we perform additional experiments by doubling and halving the number of residual blocks and the number of filters in each convolutional layer.
As it can be seen, the \DECO{} architecture is quite robust but can be potentially finetuned to each target dataset to further increase performance.
In table \ref{table:DECO_speed} we report runtimes for the considered networks. As the results show, while \DECO{} requires some extra computation time, in real life this is actually offset  by the fact that only $\frac{1}{3}$ of the data is being moved to the GPU.
\begin{table}
\centering
$\begin{array}{|l|c|}
\hline
\text{Network} & \text{Time (ms)} \\
\hline
\text{CaffeNet} & 695 \\
\hline
\text{VGG} &  1335 \\
\hline
\text{GoogleNet} &  1610 \\
\hline
\text{ResNet-50} &  1078\\
\hline
\text{\DECO{} colorizer} & 400 \\
\hline
\end{array}$
$\begin{array}{|l|c|}
\hline
\text{Network} & \text{Time (s)} \\
\hline
\text{CaffeNet} & 1.87 \\
\hline
\text{\DECO{} + CaffeNet} &  1.23 \\
\hline
\text{VGG} &  2.91 \\
\hline
\text{\DECO{} + VGG} &  2.16\\
\hline
\end{array}$
\caption{Left: forward-backward time for 50 iterations, as by \textit{caffe time}. Right:  feature extraction times for 100 images; note that using \DECO{} actually speeds up the procedure. 
We explain this by noting that \DECO{} uses single channel images and thus needs to transfer only $\frac{1}{3}$ of the data from memory to the GPU - clearly the bottleneck here. }
\label{table:DECO_speed}
\vspace{-0.5cm}
\end{table}

\begin{table*}[!htb]
\centering
$\begin{array}{|l|c|c|c|}
\hline
\text{Method:}                                      & \text{Washington\cite{washington}}          & \text{JHUIT-50\cite{jhuit}}    & \text{BigBIRD Reduced\cite{singh2014bigbird}}        \\
\hline
\text{VGG16 on Grayscale}	& 74.9 & 33.7 & 22 \\
\text{VGG16 on ColorJet}	& 75.2 & 35.3 & 19.9 \\
\text{VGG16 on SurfaceNormals}	& 75.3 & 30.8 & 16.8 \\
\text{VGG16 on SurfaceNormals++}	& 77.3 & 35.8 & 11.5 \\
\text{VGG16 \DECO{} learned on Washington} & \mathbf{79.6} & \mathbf{52.7} & 22.8 \\
\text{VGG16 \DECO{} learned on JHUIT-50} 	& 78.1	& 51.2 & 23.7 \\
\hline
\text{CaffeNet on Grayscale}	& 76.6 & 44.6 & 22.9 \\
\text{CaffeNet on ColorJet}	& 78.8 & 45.0 & 22.7 \\
\text{CaffeNet on SurfaceNormals}	& 79.3 & 38.3 & 18.9 \\
\text{CaffeNet on SurfaceNormals++}	& 81.4 & 44.8 & 14.0 \\
\text{CaffeNet \DECO{} learned on Washington} & \textcolor{red}{\mathbf{83.1}} & 53.1 & \textcolor{red}{\mathbf{28.6}} \\
\text{CaffeNet \DECO{} learned on JHUIT-50} 	& 79.1	& \textcolor{red}{\mathbf{57.5}} & 25.2 \\
\hline
\hline
\text{GoogleNet on ColorJet}	& 73.5 & 40.0 & 21.8 \\
\text{GoogleNet on SurfaceNormals}	& 72.9 & 36.5 & 18.4 \\
\text{GoogleNet on SurfaceNormals++}	&\mathbf{76.7} & 41.5 & 13.9 \\
\text{GoogleNet \DECO{} learned on Washington} & - & \mathbf{51.9} & \mathbf{25.2} \\
\text{GoogleNet \DECO{} learned on JHUIT-50} 	& 76.6 & - & 24.4 \\
\hline
\text{ResNet50 on ColorJet}	& 75.1 & 38.9 & 18.7 \\
\text{ResNet50 on SurfaceNormals}	& 77.4 & 33.2 & 16.5 \\
\text{ResNet50 on SurfaceNormals++}	& \mathbf{79.6} & 45.4 & 13.8 \\
\text{ResNet50 \DECO{} learned on Washington} & - & \mathbf{45.5} & 23.9 \\
\text{ResNet50 \DECO{} learned on JHUIT-50} 	& 76.4 & - & \mathbf{24.7}\\
\hline

\end{array}$
\caption{Object classification experiments in the depth domain,  comparing \DECO{} and hand crafted mappings, using $5$ pre-trained networks 
as feature extractors. Best results for each network-dataset combination are in \textbf{bold}, overall best in \textcolor{red}{\textbf{red bold}}. Extensive experiments were performed on VGG and Caffenet, while GoogleNet and ResNet act as reference.}
\label{table:comparison}
\vspace{-0.2cm}
\end{table*}

\begin{table}[!htb]
\centering
$\begin{array}{|l|c|c|c|}
\hline
\text{filters/blocks } & \text{4 blocks} & \text{8 blocks}   & \text{
  16 blocks}  \\
\hline
\text{32 filters} & 56.5 & 52.8 & 57.1  \\
\hline
\text{64 filters } & 56.8 & 53.1 & 53.6  \\
\hline
\text{128 filters } & 53.1 & 53.9 & 53.3 \\
\hline
\end{array}$
\caption{\DECO{} ablation study, learned on Washington, tested on JHUIT-50. Grid search optimization over width and depth of generator architecture shows improved results. }
\label{table:DECO_ablation}
\vspace{-0.5cm}
\end{table}

\subsection{Finetuning}
\label{fine-tuning}
In our finetuning experiments we focused on the best performing network from the ablation, the CaffeNet (which is also used by current competitors \cite{eitel2015multimodal,aakerberg2017depth}), to see up to which degree the network could exploit a given mapping. 
The protocol was quite simple: all layers were free to move equally, the starting learning rate was $0.001$ (with step down after $45\%$) and the solver was \textit{SGD}. Training went on for $90$ epochs for the Washington and JHUIT-50 datasets and $30$ eps. for BigBIRD (a longer training was detrimental for all settings). 
To ensure a fair comparison with the static mapping methods, the \DECO{} part of the network was kept frozen during finetuning. 

Results are reported in Table \ref{table:finetuning_comparison}. We see that here 
the gap between hand-crafted and learned colorization methods is reduced (very likely the network is compensating existing weaknesses). 
\textit{SurfaceNormals++} performs pretty well on Washington, but less so on the other two datasets (it's actually the worse on BigBIRD). Surprisingly, the simple grayscale conversion is the one that performs best on BigBIRD, but lags clearly behind on all other settings.
\DECO{} on the other hand, performs comparably to the best mapping on every single setting \textbf{and} has a $5.9\%$ lead on JHUIT-50; we argue that it is always possible to find a shallow mapping that performs very well on a specific dataset, but there are no guarantees it can generalize.


\subsection{RGB-D}
\label{rgb-d}
While this paper focuses on how to best perform recognition in  the depth modality using convnets, we wanted to provide a reference value for RGB-D object classification using \DECO\ on the depth channel. To classify RGB images we follow \cite{aakerberg2017depth} and use a pretrained VGG16 which we finetuned on the target dataset (using the previously defined protocol).
RGB-D classification is then performed, without further learning, by computing the weighted average (weight factor $\alpha$ was cross-validated) of the \textit{fc8} layers from the RGB and Depth networks and simply selecting the most likely class (the one with the highest activations). This cue integration scheme can be seen as one of the simplest, off-the-shelf algorithm for doing classifications using two different modalities \cite{TommasiOC08}. We excluded BigBIRD from this setting, due to lack of competing works to compare with. 

Results are reported in Tables \ref{table:washington_comparison}-\ref{table:jhuit_comparison}.  We see that \DECO{} produces results on par or slightly superior to the current state of the art, even while using an extremely simple feature fusion method. This is remarkable, as competitors like 
\cite{aakerberg2017depth,eitel2015multimodal}
use instead sophisticated, deep learning based cue integration methods. Hence, our ability to compete in this setting is all due to the \DECO\ colorization mapping, clearly more powerful than the other baselines.  
It is worth stressing that, in spite of the different cue integration and depth mapping approaches compared in Tables \ref{table:washington_comparison}-\ref{table:jhuit_comparison}, convnet results on RGB are already very high, hence in this setting the advantage brought by a superior performance on the depth channel tends to be covered. Still, on Washington we achieve the second best result, and on JHUIT-50 we get the new state of the art.

\begin{table*}[!htb]
\centering
$\begin{array}{|l|c|c|c|}
\hline
\text{Method:}                                      & \text{Washington\cite{washington}}          & \text{JHUIT-50\cite{jhuit}}    & \text{BigBIRD Reduced\cite{singh2014bigbird}}        \\
\hline
\text{CaffeNet on Grayscale}	& 82.7 \pm 2.1 & 53.7 & \mathbf{29.6} \\
\text{CaffeNet on ColorJet}	& 83.8 \pm 2.7 & 54.1 & 25.4 \\
\text{CaffeNet on SurfaceNormals++}	& \mathbf{84.5} \pm 2.9 & 55.9 & 17.0 \\
\hline
\text{CaffeNet \DECO{} learned on Washington} & 84.0 \pm 2.0 & 60.0 & - \\
\text{CaffeNet \DECO{} learned on JHUIT-50} 	& 82.3 \pm 2.3	& \mathbf{62.0} & - \\
\text{CaffeNet \DECO{} learned on Washington + JHUIT-50} 	& 84.0 \pm 2.3	& 61.8	& 28.0\\
\hline
\end{array}$
\caption{CaffeNet finetuning using different colorization techniques.}
\label{table:finetuning_comparison}
\vspace{-0.5cm}
\end{table*}

\begin{table}
\centering
$\begin{array}{|l|c|c|c|}
\hline
\text{Method:}                                      & \text{RGB}          & \text{Depth}    & \text{RGB-D}        \\
\hline
\text{FusionNet\cite{eitel2015multimodal}}	& 84.1 \pm 2.7 & 83.8 \pm 2.7 & 91.3 \pm 1.4 \\
\text{CNN + Fisher \cite{li2015hybrid}}	& \mathbf{90.8} \pm 1.6 & 81.8 \pm 2.4 & \mathbf{93.8} \pm 0.9 \\
\text{DepthNet \cite{carlucci2016deep}} & 88.4 \pm 1.8 & 83.8 \pm 2.0 & 92.2 \pm 1.3 \\
\text{CIMDL \cite{wang2016correlated}} 	& 87.3 \pm 1.6	& 84.2 \pm 1.7 & 92.4 \pm 1.8 \\
\text{FusionNet enhanced\cite{aakerberg2017depth}} 	& 89.5 \pm 1.9	& \mathbf{84.5} \pm 2.9	& 93.5 \pm 1.1\\
\hline
\text{\DECO{}} 	& 89.5 \pm 1.6	& 84.0 \pm 2.3	& 93.6 \pm 0.9\\
\hline
\end{array}$
\caption{Selected results on Washington RGB-D}
\label{table:washington_comparison}
\vspace{-0.2cm}
\end{table}

\begin{table}[!htb]
\centering
$\begin{array}{|l|c|c|c|}
\hline
\text{Method:}                                      & \text{RGB}          & \text{Depth}    & \text{RGB-D}        \\
\hline
\text{DepthNet \cite{carlucci2016deep}} & 88.0 & 55.0 & 90.3 \\
\text{Beyond Pooling \cite{jhuit}} 	& -	& - & 91.2 \\
\text{FusionNet enhanced\cite{aakerberg2017depth}} 	& \mathbf{94.7}	& 56.0	& 95.3\\
\hline
\text{\DECO{}} 	& \mathbf{94.7}	& \mathbf{61.8}	& \mathbf{95.7}\\
\hline
\end{array}$
\caption{Selected results on JHUIT-50}
\label{table:jhuit_comparison}
\vspace{-0.5cm}
\end{table}



\label{expers}

\section{CONCLUSIONS}

This paper presented a network for learning deep colorization mappings. Our architecture follows the residual philosophy, learning how to map depth data to RGB images for a given pre-trained convolutional neural network. By using our \DECO\ algorithm, as opposed to the hand-crafted colorization mappings commonly used in the literature, we obtained a significant jump in performance over three different benchmark databases, using four different popular deep networks pre trained over ImageNet. The visualization of the obtained colorized images further confirms how our algorithm is able to capture the rich informative content and the different facets of depth data. 
All the deep depth mappings presented in this paper are available at \url{https://github.com/fmcarlucci/de2co}.
Future work will further investigate the effectiveness and generality of our approach, testing it on other RGB-D classification and detection problems, with various fine-tuning strategies and on several deep networks, pre-trained over different RGB databases, and in combination with RGB convnet with more advanced multimodal fusion approaches.

\bibliography{biblio} 
\bibliographystyle{IEEEtran}

\end{document}